\documentclass[12pt]{article}
\usepackage[dvips]{color,graphicx}
\usepackage{amsmath,amssymb}
\usepackage{epsf}

\makeatletter
\newcommand{\figcaption}[1]{\def\@captype{figure}\caption{#1}}
\newcommand{\tblcaption}[1]{\def\@captype{table}\caption{#1}}
\makeatother

\newtheorem{theorem}{Theorem}
\newtheorem{lemma}[theorem]{Lemma}
\newtheorem{corollary}[theorem]{Corollary}
\newtheorem{example}[theorem]{Example}

\title{Bayesian Estimation of Multidimensional Latent Variables and Its Asymptotic Accuracy}

\author{Keisuke Yamazaki\\
       k.yamazaki@aist.go.jp \\
       Artificial Intelligence Research Center,\\
       National Institute of Advanced Industrial Science and Technology\\
       2-3-26 Aomi, Koto-ku, Tokyo, Japan
	}
\date{}
\begin{document}
\sloppy
\maketitle

\begin{abstract}
Hierarchical learning models, such as
mixture models and Bayesian networks,
are widely employed for unsupervised learning tasks, such as clustering analysis.
They consist of observable and latent variables,
which represent the given data and their underlying generation process, respectively.
It has been pointed out that conventional statistical analysis is not applicable
to these models, because redundancy of the latent variable produces singularities in the parameter space.
In recent years, a method based on algebraic geometry
has allowed us to analyze the accuracy of predicting observable variables when using Bayesian estimation.
However, how to analyze latent variables has not been sufficiently studied, even
though one of the main issues in unsupervised learning
is to determine how accurately the latent variable is estimated.
A previous study  proposed a method that can be used
when the range of the latent variable is redundant
compared with the model generating data.
The present paper extends that method to the situation in which the latent variables have 
redundant dimensions.
We formulate new error functions
and derive their asymptotic forms.
Calculation of the error functions is demonstrated in two-layered
Bayesian networks.
\newline
{\bf Keywords:}
  unsupervised learning, hierarchical parametric models, Bayesian statistics, algebraic geometry, singularities
\end{abstract}

\section{Introduction}
\sloppy
Hierarchical parametric models, such as Gaussian mixtures, Boltzmann machines,
and Bayesian networks, are often used for unsupervised learning.
These models use two variables to express their structure:
observable variables that represent the given data,
and latent variables that imply the hidden structure or labels.
The task of unsupervised learning is to estimate the latent variables.
For example, in cluster analysis with a Gaussian mixture model,
the latent variable expresses a cluster label that indicates which cluster generated the data point,
and the task is to assign a label to each of the data points.

Due to their structure, hierarchical models have singularities
in their parameter space.
Probabilistic learning models generally fall into one of two classes: regular and singular.
A regular model has a one-to-one relation
between the parameters and the expression of the model as a probability function; a singular model does not.
Singularities adversely affect conventional statistical analysis; they reduce the rank of the Fisher information matrix, which is required by the analysis to be positive definite.
Thus, results based on regularity are not valid in singular models; this includes asymptotic analysis on the maximum likelihood estimator
and the use of model selection criterion, such as the BIC or MDL \cite{Schwarz,Rissanen}.

To tackle this issue,
modern mathematics, such as the field of algebraic geometry, has been applied to the analysis of models
using Bayesian statistics \cite{Watanabe01a,Watanabe09:book}.
Asymptotic properties of some important functions have been clarified
by using this method.
As a substitute for the BIC, the marginal likelihood
has been used to evaluate various models \cite{Yamazaki03a,Yamazaki03b,Yamazaki05c,Rusakov,Aoyagi05,Aoyagi10,Zwiernik11}.
The results directly reveal the accuracy of measuring the generalization error when predicting observable variables.
A novel model selection criterion based on the asymptotic error has been proposed \cite{Watanabe10:WAIC}.
The posterior distribution plays an important role in the Bayesian statistics.
Specifically, its convergence rate is one of the main concerns in the statistical literature,
and has been analyzed (e.g., \cite{Ghosal+2000,LeCam1973,Ibragimov+1981}).
For the cases with singularities, the rate based on the Wasserstein metrics is elucidated in both finite and infinite
mixture models \cite{Nguyen2013}.
Using the algebraic techniques, the identifiability of the hierarchical models is discussed in \cite{Allman+2009}.

The estimation accuracy of the latent variables has not been sufficiently studied, even
though they are one of the main concerns in unsupervised learning.
It is not straightforward to evaluate the estimation result,
because the optimality varies due to many factors,
such as the intended use of the results and prior knowledge about the given data.
In the case of hierarchical models, however,
we can estimate the variables and evaluate the results in a distribution-based manner;
the probability of the latent variables is naturally formulated from the model expression,
and the accuracy can be defined by differences in the distributions.
Using this approach, an analysis method has been proposed and the asymptotic form of an error function
is available for applying the maximum likelihood method and the Bayes method to regular cases \cite{Yamazaki14a}.
The algebraic geometrical method is again applicable to singular cases in the Bayes method.
It has been found that singularities play an essential role in determining the convergence of the error \cite{Yamazaki15a}.

Singularities appear when the model is redundant compared with
the true distribution, from which the data are generated.
There are two cases in which redundancy emerges: the latent value may have a 
redundant range or it may have redundant dimensions.
In our previous studies \cite{Yamazaki14a,Yamazaki15a},
we considered mixture models, in which there is only one latent variable
that expresses to which component the data point belongs.
For example, suppose a model has $K$ components, and the latent variable is described by $y\in \{1,\dots, K\}$,
while the true distribution has $K^*$ components, where $K^*<K$;
this is a case where the range has redundancy.
In the present paper,
we will consider the other case, that is, where the latent variable is multidimensional, and it has redundancy.
For example, in a Bayesian network,
let the latent nodes be $y=(y_1,y_2,\dots,y_K)$.
Note that each element has a range, that is, $y_i\in \{1,\dots,L_i\}$.
When the true model has the latent nodes $y=(y_1,\dots,y_{K^*})$ for $K^*<K$,
singularities appear in the parameter space of the model.
The standard error function measures the accuracy with the Kullback-Leibler (KL) divergence;
note that the following loss function, which was used in our previous study \cite{Yamazaki14a}, is not well-defined:
\begin{align}
\sum_y q(y|x) \ln \frac{q(y|x)}{p(y|x)},
\end{align}
where $x$ is the observable variable, and $q(y|x)$ and $p(y|x)$ are, respectively, the true and estimated distributions
of the latent variable.
Since the latent variable $y$ will have different dimensions in $q$ and $p$,
this loss function cannot be used when the latent variable is multidimensional.

In the present paper, we provide new definitions for the error function for the multidimensional case
and derive the asymptotic form of their Bayesian estimation.
Section 2 briefly summarizes the Bayesian estimation of the latent variable.
Section \ref{sec:2EF} introduces some estimation tasks and the definitions of the error
for both the non-redundant and the redundant case.
Section \ref{sec:analysis} presents the asymptotic forms,
and Section \ref{sec:app2BN} demonstrates the actual error that occurs with two-layered Bayesian networks.
A discussion and our conclusions are presented in Sections \ref{sec:Dis} and \ref{sec:Conc}, respectively.
\section{Bayesian Estimation of a Latent Variable }
\label{sec:BLVE}
This section formulates the Bayesian estimation of a latent variable.
Let a learning model be expressed as
\begin{align}
p(x|w) =& \sum_y p(x,y|w),\label{eq:modelexp}
\end{align}
where $x\in R^M$ is an observable variable describing $M$-dimensional data,
$w$ stands for the parameters, and
$y \in \{1,\ldots,L\}^K$ is a latent variable describing $K$-dimensional discrete labels.
By replacing $\int dx$ with $\sum_x$,
the results in the present paper still hold when the observable variable is discrete.

We now show some examples of hierarchical models.
A Gaussian mixture model is given by
\begin{align}
p(x|w) =& a_1 \mathcal{N}(x|b_1) + a_2\mathcal{N}(x|b_2),\label{eq:GM}
\end{align}
where $\mathcal{N}(x|b)$ is a Gaussian distribution with the parameter $b$.
Note that the mean and the variance parameters can be included in $b$.
The parameter of this model is $w=\{a_1,b_1,b_2\}$, where $0\le a_1 \le 1$ and $a_2=1-a_1$.
The mixture has a one-dimensional label $y=\{1,2\}$ that describes the number of components,
and the model expression of Eq.~(\ref{eq:modelexp}) is rewritten as
\begin{align}
p(x|w) = \sum_{y=1,2} a_y \mathcal{N}(x|b_y) = \sum_{y=1,2} p(x,y|w).
\end{align}
\begin{figure}[t]
\begin{minipage}[c]{.3\textwidth}
\includegraphics[angle=-90, width=\textwidth]{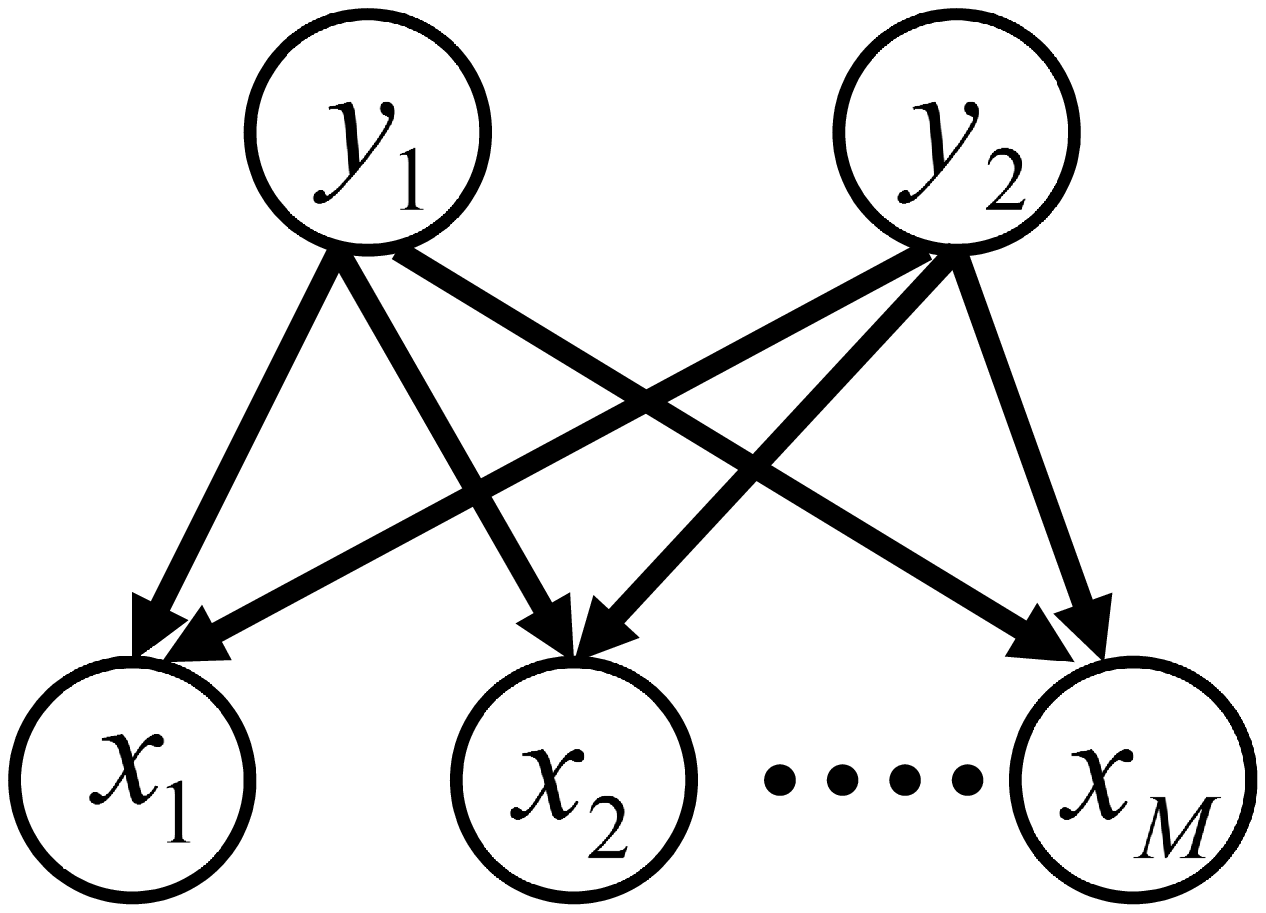}
\end{minipage}
\def\@captype{table}
\begin{minipage}[c]{.3\textwidth}
\begin{tabular}{c||c|c}
$y_1$ & 0 & 1\\
\hline
& $a_{10}$ & $1-a_{10}$
\end{tabular}
\\
\begin{tabular}{c}
\\
\end{tabular}
\\
\begin{tabular}{c||c|c}
$y_2$ & 0 & 1 \\
\hline
& $a_{20}$ & $1-a_{20}$
\end{tabular}
\end{minipage}
\begin{minipage}[c]{.3\textwidth}
\begin{tabular}{c||c|c|c|c}
$x_m$ & $y_1$ & $y_2$ & 0 & 1 \\
\hline
& 0 & 0 & $b_{00m}$ & $1-b_{00m}$ \\
& 0 & 1 & $b_{01m}$ & $1-b_{01m}$ \\
& 1 & 0 & $b_{10m}$ & $1-b_{10m}$ \\
& 1 & 1 & $b_{11m}$ & $1-b_{11m}$
\end{tabular}
\end{minipage}
\caption{Simple two-layer Bayesian network: structure (left panel), CPTs of nodes $y_1$ and $y_2$ (middle panel),
and CPT of node $x_m$ (right panel).}
\label{fig:simpleBN}
\end{figure}
Next, we introduce an example with a multidimensional label;
the simple two-layered binary Bayesian network described in Fig. \ref{fig:simpleBN} is given by
\begin{align}
p(x|w) =& \sum_{y_1=0,1}\sum_{y_2=0,1} a_{1y_1}a_{2y_2}\prod_{m=1}^M b_{y_1y_2m}^{(1-x_m)}(1-b_{y_1y_2m})^{x_m}, \label{eq:def_p_simpleBN}
\end{align}
where $x=(x_1,\dots,x_M) \in \{0,1\}^M$, and the parameter is $w=\{a_{10},a_{20},b_{ijm}\}$ for $i,j=0,1$ and $m=1,\dots,M$,
such that $0\le a_{10}\le 1$, $a_{11}=1-a_{10}$, $0\le a_{20}\le 1$, $a_{21}=1-a_{20}$,
and $0\le b_{ijm}\le 1$.
This model has a two-dimensional label $y=(y_1,y_2) \in \{0,1\}^2$.
We will show a network with a general-dimensional label in Section \ref{sec:app2BN}.
It is easily confirmed that the two-layered Bayesian network is a natural extension of the naive Bayesian network,
which has a single latent latent node.
Since there are more than one latent node, the two-layered networks can deal with the multilabel clustering,
where the effect of multiple latent factors on the observable variables is reflected in the result 
and the necessary and the nuisance factors are distinguished in the unsupervised manner.

Let the data-generating model, referred to as the true model, be expressed as
\begin{align}
q(x) = \sum_y q(x,y),
\end{align}
where the latent variable is described by $y\in \{1,\ldots,L^*\}^{K^*}$.
The difference in the dimensions, such that $K > K^*$, implies the existence of redundant latent variables.
The present paper assumes that the learning model can attain the true model,
i.e., $K\ge K^*$, $L \ge L^*$, and that the set $W_t=\{w^*: p(x|w^*)=q(x)\}$ is not empty.
Note that the range of each dimension of $y$ can be different; for example, it may be that
$y=(y_1,\dots,y_{K^*}), y_k\in \{1,\dots,L^*_k\}$.
For simplicity, we consider the case $L^*_1=\dots=L^*_{K^*}$.
In the present paper, we will formulate error functions for the estimation of the latent variable, both with and without redundant dimensions.
Since the error function will based on the KL divergence including the factor $\ln q(y|x) / p(y|x)$,
the situation $K<K^*$ or $L<L^*$ is not well-defined;
the factor diverges to infinity when $p(y|x)=0$ and $q(y|x)\ne 0$.
This is the reason why we consider $K\ge K^*$ and $L\ge L^*$.

Let us formalize the data set.
The true model is assumed to generate data that are independently and identically distributed.
Let a set of $n$ given data be $X^n=\{x_1,\ldots,x_n\}$.
The corresponding latent variables are denoted $Y^n=\{y_1,\ldots,y_n\}$.
The detailed expressions are $x_i=(x_{i1},\dots,x_{iM})$ and $y_i = (y_{i1},\dots,y_{iL})$, respectively.
We will write them as pairs as follows: $(X^n,Y^n)=\{(x_1,y_1),\ldots,(x_n,y_n)\}$.

To deal with the redundant dimensions of the latent variable,
we divide the variable in the learning model into two parts:
$y_i=(y^{(1)}_i,y^{(2)}_i)$, where $y^{(1)}_i\in \{1,\ldots,L\}^{K^*}$,
and $y^{(2)}_i\in \{1,\ldots,L\}^{K-K^*}$.
More precisely, for $y_i=(y_{i1},\dots,y_{iK})$,
the two parts are given by
\begin{align}
y^{(1)}_i &= (y^{(1)}_{i1},\dots,y^{(1)}_{iK^*}) = (y_{i1},\dots,y_{iK^*}),\\
y^{(2)}_i &= (y^{(2)}_{i1},\dots,y^{(2)}_{iK-K^*}) = (y_{iK^*+1},\dots,y_{iK}).
\end{align}
We also use $Y^n_i=\{y_1^{(i)},\ldots,y_n^{(i)}\}$ for $i=1,2$.


We consider a distribution-based estimation of the latent variables; that is,
estimating $Y^n$ from the given data $X^n$ is represented by constructing the distribution $p(Y^n|X^n)$.
In general Bayesian statistics, the marginal likelihood is defined by
\begin{align}
Z(X^n) =& \int \prod_{i=1}^n p(x_i|w)\varphi(w;\eta)dw,
\end{align}
where $\varphi(w;\eta)$ is a prior with a hyperparameter $\eta$.
We define the marginal likelihood of the complete data as
\begin{align}
Z(X^n,Y^n) =& \int \prod_{i=1}^n p(x_i,y_i|w)\varphi(w;\eta)dw,
\end{align}
where $Z(X^n)=\sum_{Y^n}Z(X^n,Y^n)$.
As introduced in \cite{Yamazaki14a}, the Bayesian estimation of the latent variables is given by
\begin{align}
p(Y^n|X^n) =& \frac{Z(X^n,Y^n)}{Z(X^n)}.
\end{align}
The true distribution is given by
\begin{align}
q(Y^n|X^n) = \prod_{i=1}^n\frac{q(x_i,y_i)}{q(x_i)}= \prod_{i=1}^n q(y_i|x_i).
\end{align}
Note that $Y^n\in \{1,\ldots,L\}^{Kn}$ in $p(Y^n|X^n)$,
while $Y^n \in \{1,\ldots,L^*\}^{K^*n}$ in $q(Y^n|X^n)$.
To avoid this confusion, we use the following summation symbols:
\begin{align}
\sum_{Y^n:K}^L &=\sum_{y_{11}=1}^L\!\cdots\!\sum_{y_{1K}=1}^L\sum_{y_{21}=1}^L\!\cdots\!\sum_{y_{2K}=1}^L\!\cdots\!\sum_{y_{n1}=1}^L\!\cdots\!\sum_{y_{nK}=1}^L,\\
\sum_{Y^n:K^*}^{L^*}\!\! &=\!\!
\sum_{y_{11}=1}^{L^*}\!\!\!\cdots\!\!\!\sum_{y_{1K^*}=1}^{L^*}\sum_{y_{21}=1}^{L^*}\!\!\!\cdots\!\!\!
\sum_{y_{2K^*}=1}^{L^*}\!\!\!\cdots\!\!\!\sum_{y_{n1}=1}^{L^*}\!\!\!\cdots\!\!\!\sum_{y_{nK^*}=1}^{L^*}.
\end{align}
\section{Error Functions for Multidimensional Latent Variables}
\label{sec:2EF}
In this section, we propose and formulate error functions for measuring the estimation accuracy.
We consider two cases: when the latent variable does not have redundant dimensions, $K=K^*$; and when it does, $K>K^*$.
\subsection{Non-Redundant Case}
If there are no redundant dimensions in the latent variables, i.e., $K=K^*$,
the error function is given by the KL divergence from the true distribution
to the estimated distribution \cite{Yamazaki11c}:
\begin{align}
D_{n1}(n) = \frac{1}{n}E_{X^n}\bigg[ \sum_{Y^n:K^*}^{L^*} q(Y^n|X^n)\ln\frac{q(Y^n|X^n)}{p(Y^n|X^n)} \bigg], \label{eq:def_Dn1}
\end{align}
where $E_{X^n}[f(X^n)]=\int f(X^n) q(X^n)dX^n$.
This is a natural extension of a one-dimensional latent variable.
We refer to this as a joint error function,
since it corresponds to measuring the joint probability of the dimensions,
i.e., $q(Y^n|X^n)=\prod_{i=1}^n q(y_{i1},\dots,y_{iK^*}|x_i)$.

We also consider the situation in which there are partial target dimensions to be estimated.
Let us divide $y^{(1)}_i$ into two parts, as follows:
\begin{align}
y^{(1)}_i =& (y^{(11)}_i,y^{(12)}_i),\\
y^{(11)}_i =& (y_{i1},\dots,y_{iK_t}),\\
y^{(12)}_i =& (y_{iK_t+1},\dots,y_{iK^*}),
\end{align}
where $k=1,\dots,K_t$ such that $K_t<K^*$ is the dimensionality of the target.
We use the notation $Y^n_{1j}=\{y^{(1j)}_1,\dots,y^{(1j)}_n\}$ for $j=1,2$.
In this case, the nuisance dimensions $k=K_t+1,\dots,K^*$ will be ignored when evaluating the accuracy.
The error function is given by
\begin{align}
D_{n2}(n) =& \frac{1}{n}E_{X^n}\bigg[ \sum_{Y^n_{11}:K^*}^{L^*} q(Y^n_{11}|X^n) \ln \frac{q(Y^n_{11}|X^n)}{p(Y^n_{11}|X^n)}\bigg],
\end{align}
where the probability of $Y^n_{11}$ means that the nuisance dimensions $Y^n_{12}$ are marginalized out:
\begin{align}
q(Y^n_{11}|X^n) =& \sum_{Y^n_{12}:K^*}^{L^*} q(Y^n|X^n),\\
p(Y^n_{11}|X^n) =& \sum_{Y^n_{12}:K^*}^{L^*} p(Y^n|X^n);
\end{align}
and we use the following summation symbols:
\begin{align}
\sum_{Y^n_{11}:K^*}^{L^*} =& \sum_{y_{11}=1}^{L^*}\dots \sum_{y_{1K_t}=1}^{L^*} \sum_{y_{21}=1}^{L^*}\dots \sum_{y_{2K_t}=1}^{L^*}\dots \sum_{y_{n1}=1}^{L^*}\dots \sum_{y_{nK_t}=1}^{L^*},\\
\sum_{Y^n_{12}:K^*}^{L^*} =& \sum_{y_{1K_t+1}=1}^{L^*}\dots \sum_{y_{1K^*}=1}^{L^*}\dots \sum_{y_{nK_t+1}=1}^{L^*}\dots \sum_{y_{nK^*}=1}^{L^*}.
\end{align}
We refer to $D_{n2}(n)$ as the marginal type since the irrelevant dimensions are marginalized out.
In the practical situations, the partial target dimensions are not often distinguishable from the irrelevant ones
since the estimated labels have symmetry.
The present paper considers the marginal type error to investigate the relation
between the number of the target dimensions and the accuracy from the theoretical perspective.
The label symmetry will be discussed in Section \ref{sec:Dis_symmetry}.

It is easy to prove the following lemma, which holds for any arbitrary number of data points $n$.
\begin{lemma}
\label{lem:Dn1geDn2}
The joint-type error is larger than the marginal-type error:
\begin{align}
D_{n1}(n) \ge D_{n2}(n).
\end{align}
\end{lemma}
{\bf (Proof):}
Based on the log-sum inequality on $\sum_{Y^n_{12}:K^*}^{L^*}$, we immediately obtain that
\begin{align}
D_{n2}(n) =& \frac{1}{n}E_{X^n}\bigg[ \sum_{Y^n_{11}:K^*}^{L^*} \# Y^n_{12} \bigg\{ \sum_{Y^n_{12}:K^*}^{L^*}\frac{1}{\# Y^n_{12}}q(Y^n|X^n)\bigg\} 
\ln \frac{\sum_{Y^n_{12}:K^*}^{L^*}\frac{1}{\# Y^n_{12}}q(Y^n|X^n)}
{\sum_{Y^n_{12}:K^*}^{L^*}\frac{1}{\# Y^n_{12}}p(Y^n|X^n)}\bigg]\\
\le& \frac{1}{n}E_{X^n}\bigg[ \sum_{Y^n_{11}:K^*}^{L^*} \# Y^n_{12} \sum_{Y^n_{12}:K^*}^{L^*} \frac{1}{\# Y^n_{12}} q(Y^n|X^n) 
\ln \frac{\frac{1}{\# Y^n_{12}} q(Y^n|X^n)}{\frac{1}{\# Y^n_{12}} p(Y^n|X^n)}\bigg]\\
=& D_{n1}(n),
\end{align}
where $\# Y^n_{12}=L^{*(K^*-K_t)}$. {\bf (End of Proof)}

This lemma shows the relation between the estimation target and the accuracy:
the lower the dimensions of the target, the smaller the error.
\begin{example}
\label{ex:mag_tar_LV}
Let us consider two different target dimensions $K_t=K_A$ and $K_t=K_B$, where $K_A<K_B$. 
Define the error functions $D_{n2A}(n)$ and $D_{n2B}(n)$ in the marginal-type manner for these two dimensions, respectively.
Using the same way of the proof of Lemma \ref{lem:Dn1geDn2}, we can easily obtain the following relation,
\begin{align*}
D_{n2A}(n) < D_{n2B}(n).
\end{align*}
\end{example}
\subsection{Redundant Case}
We now consider the redundant case, i.e., $K>K^*$.
The definition in Eq.~(\ref{eq:def_Dn1}) is not applicable to this case,
since the domain spaces of $q(Y^n|X^n)$ and $p(Y^n|X^n)$ are different.
In the expression for the latent variable,
the true and estimated distributions are described by
$q(Y^n_1|X^n)$ and $p(Y^n_1,Y^n_2|X^n)$, respectively.
These notations clarify that the KL divergence of Eq.~(\ref{eq:def_Dn1}) is not well-defined in this case.

To measure the distribution on $Y^n_1$,
we can define the following two error functions:
\begin{align}
D_{r1}(n) =& \frac{1}{n}E_{X^n}\bigg[ \sum_{Y^n_1:K^*}^{L^*} q(Y^n_1|X^n)\ln\frac{q(Y^n_1|X^n)}{p(Y^n_1,Y^n_2=1|X^n)}\bigg],\\
D_{r2}(n) =& \frac{1}{n}E_{X^n}\bigg[ \sum_{Y^n_1:K^*}^{L^*} \!\! q(Y^n_1|X^n)\ln\frac{q(Y^n_1|X^n)}{\sum_{Y^n_2:K-K^*}^L p(Y^n_1,Y^n_2|X^n)}\bigg],
\end{align}
where $p(Y^n_1,Y^n_2=1|X^n)$ means that all elements of $Y^n_2$ are unity.
The first error, $D_{r1}(n)$, can be regarded as a joint-type error
when the values in $Y^n_2$ are fixed.
The second is a marginal-type error;
the target dimensionality is $Y^n_1$, and the nuisance dimensionality is $Y^n_2$.

These two errors use different definitions for the estimated distribution on $Y^n_1$.
In the marginal-type error, the redundant dimensions $Y^n_2$ are marginalized out.
This means that the error only evaluates the target dimensions $Y_1^n$,
even if probabilities have been assigned in $Y^n_2$.
In the joint-type error, the estimation is expected to construct
the probabilities only on $Y^n_1$, even though the latent variable has redundant dimensions;
the essential dimensions must be detected successfully,
and the redundant ones must be assigned a fixed value that does not affect the result.
In general, latent variables are equivalent to nought
if their values are deterministically assigned.
For simplicity,
the definition selects $Y^n_2=1$ as an example.
The joint-type error evaluates the effect that the redundant dimensions have on
the estimation result.

From the above definitions, we have the following lemma,
which holds for arbitrary $n$.
\begin{lemma}
\label{lem:Dr1geDr2}
The magnitudes of the errors are related as follows:
\begin{align}
D_{r1}(n) \ge D_{r2}(n).
\end{align}
\end{lemma}
{\bf (Proof):}
We have
\begin{align}
\sum_{Y^n_2:K-K^*}^L p(Y^n_1,Y^n_2|X^n) \ge p(Y^n_1,Y^n_2=1|X^n),
\end{align}
which directly shows that $D_{r1}(n) \ge D_{r2}(n)$. {\bf (End of Proof)}

When the target dimension is restricted to $Y^n_{11}$ in the similar way to $D_{n2}(n)$,
the nuisance dimensions consist of $Y^n_{12}$ and $Y^n_2$.
The error function is thus of the marginal type and is given by
\begin{align}
D_{r3}(n) =& \frac{1}{n}E_{X^n}\bigg[ \sum_{Y^n_{11}:K^*}^{L^*} q(Y^n_{11}|X^n)\ln \frac{q(Y^n_{11}|X^n)}{p(Y^n_{11}|X^n)}\bigg],
\end{align}
where the marginal probabilities are defined by
\begin{align}
q(Y^n_{11}|X^n) =& \sum_{Y^n_{12}:K^*}^{L^*} q(Y^n_1|X^n),\\
p(Y^n_{11}|X^n) =& \sum_{Y^n_{12}:K^*}^L \sum_{Y^n_2:K}^L p(Y^n_1,Y^n_2|X^n).
\end{align}

By combining Lemmas \ref{lem:Dn1geDn2} and \ref{lem:Dr1geDr2}, we obtain
the following lemma.
\begin{lemma}
The magnitudes of the errors are related as follows: 
\begin{align}
D_{r1}(n) \ge D_{r2}(n) \ge D_{r3}(n).
\end{align}
\end{lemma}
\section{Asymptotic Analysis of the Error Functions}
\label{sec:analysis}
In this section, we present the asymptotic forms of the error function for 
the non-redundant and the redundant cases.
These results are based on an extension of the methods provided in \cite{Yamazaki14a, Yamazaki15a}.
In the following Sections \ref{subsec:non-redundant} and \ref{subsec:redundant}, 
we assume $L=L^*$, which means that the range of the latent variables has no redundancy.
Then, we show their extensions to the case $L>L^*$ in Section \ref{subsec:extGenCase}.
\subsection{Free Energy Functions}
We now introduce free energy functions, which play an important role in deriving the asymptotic form of the error functions.
First, we define the following marginal likelihood functions:
\begin{align}
Z(X^n) =& \int \prod_{i=1}^n p(x_i|w)\varphi(w;\eta)dw,\\
Z(X^n,Y^n) =& \int \prod_{i=1}^n p(x_i,y_i|w)\varphi(w;\eta)dw,\\
Z_{XY_{11}}(X^n,Y^n_{11}) =& \int \prod_{i=1}^n p(x_i,y^{(11)}_i|w)\varphi(w;\eta)dw,\\
Z_{XY_1}(X^n,Y^n_1) =& \int \prod_{i=1}^n \sum_{y^{(2)}_i=1}^L 
p(x_i,y^{(1)}_i,y^{(2)}_i|w)\varphi(w;\eta)dw,\\
Z_{XY_1C}(X^n,Y^n_1) =& \int \prod_{i=1}^n p(x_i,y^{(1)}_i,y^{(2)}_i=1|w)\varphi(w;\eta)dw.
\end{align}
Note that $Z(X^n,Y^n)$ is used for only the non-redundant case,
and $p(x_i,y^{(11)}_i|w)$ in $Z_{XY_{11}}(X^n,Y^n_{11})$ has two different definitions. It is defined as
\begin{align}
p(x,y^{(11)}_i|w) =& \sum_{y^{(12)}_i=1}^L p(x,y^{(11)}_i, y^{(12)}_i|w),\\
p(x,y^{(11)}_i|w) =& \sum_{y^{(12)}_i=1}^L \sum_{y^{(2)}_i=1}^L p(x,y^{(11)}_i,y^{(12)}_i,y^{(2)}_i|w)
\end{align}
in the non-redundant and the redundant case, respectively.

The negative log marginal likelihood is referred to as the free energy.
We consider the following variants of functions of the average energy:
\begin{align}
F_X(n) =& -nS_X - E_{X^n}[\ln Z(X^n)],\\
F_{XY}(n) =& -nS_{XY} -E_{X^n,Y^n_1}[\ln Z(X^n,Y^n)], \label{eq:def_FXY}\\
F_{XY_{11}}(n) =& -nS_{XY} -E_{X^n,Y^n_1}[\ln Z_{XY_{11}}(X^n,Y^n_{11})],\\
F_{XY_1}(n) =& -nS_{XY} -E_{X^nY^n_1}[\ln Z_{XY_1}(X^n,Y^n_1)],\\
F_{XY_1C}(n) =& -nS_{XY} -E_{X^nY^n_1}[\ln Z_{XY_1C}(X^n,Y^n_1)],
\end{align}
where
\begin{align}
S_X =& -\int q(x)\ln q(x)dx,\\
S_{XY} =& -\int \sum_{y=1}^{L^*}q(x,y)\ln q(x,y)dx,
\end{align}
and the expectation is defined as
\begin{align}
E_{X^nY^n_1}[f(X^n,Y^n_1)] =& \int \sum_{Y^n_1:K^*}^{L^*} q(X^n,Y^n_1)f(X^n,Y^n_1)dX^n.
\end{align}
It is easy to find the following relations between the error functions and the free energy functions:
\begin{align}
D_{n1}(n) =& \frac{1}{n}\{ F_{XY}(n) - F_X(n) \}, \label{eq:rel_n1}\\
D_{n2}(n) =& \frac{1}{n}\{ F_{XY_{11}}(n) - F_X(n)\}, \label{eq:rel_n2}\\
D_{r1}(n) =& \frac{1}{n}\{ F_{XY_1C}(n) - F_X(n) \}, \label{eq:rel_r1}\\
D_{r2}(n) =& \frac{1}{n}\{ F_{XY_1}(n) - F_X(n) \}, \label{eq:rel_r2}\\
D_{r3}(n) =& \frac{1}{n}\{ F_{XY_{11}}(n) - F_X(n)\}. \label{eq:rel_r3}
\end{align}
For example, the equation for $D_{n1}(n)$ is derived as follows:
\begin{align}
D_{n1}(n) =& \frac{1}{n}E_{X^n}\bigg[ \sum_{Y^n:K^*}^{L^*} q(Y^n|X^n)\ln \frac{q(X^n,Y^n)}{p(X^n,Y^n)}-\ln \frac{q(X^n)}{p(X^n)}\bigg]\nonumber\\
=& \frac{1}{n}\bigg\{-nS_{XY}-E_{X^nY^n}[\ln Z(X^n,Y^n)]\bigg\}-\frac{1}{n}\bigg\{-nS_X-E_{X^n}[\ln Z(X^n)]\bigg\}\nonumber\\
=&\frac{1}{n}\big\{ F_{XY}(n) - F_X(n)\big\},
\end{align}
where $p(X^n,Y^n)$ and $p(X^n)$ correspond to $Z(X^n,Y^n)$ and $Z(X^n)$, respectively, and $q(X^n,Y^n)=q(Y^n|X^n)q(X^n)$.

Due to these relations, we will consider the asymptotic forms of the free energy functions in the following subsections.
\subsection{Asymptotic Errors in the Non-Redundant Case}
\label{subsec:non-redundant}
According to the analysis in \cite{Yamazaki14a},
the asymptotic free energy can be expressed in terms of the Fisher information matrix.
Let us here define these matrices by defining their elements, as follows:
\begin{align}
I_X(w) =& \bigg\{ E_x\bigg[\frac{\partial}{\partial w_i}\ln p(x|w) \frac{\partial}{\partial w_j}\ln p(x|w)\bigg] \bigg\}_{ij},\\
I_{XY}(w) =& \bigg\{ E_{xy}\bigg[\frac{\partial}{\partial w_i}\ln p(x,y|w) \frac{\partial}{\partial w_j}\ln p(x,y|w) \bigg]\bigg\}_{ij},\\
I_{XY_{11}}(w) =& \bigg\{ E_{xy}\bigg[\frac{\partial}{\partial w_i}\ln p(x,y^{(11)}|w) \frac{\partial}{\partial w_j}\ln p(x,y^{(11)}|w) \bigg]\bigg\}_{ij}.
\end{align}
The expectations are defined as
\begin{align}
E_x[f(x)] =& \int f(x)q(x)dx,\\
E_{xy}[f(x,y)] =& \int \sum_y f(x,y)q(x,y)dx.
\end{align}

Using these matrices, we can describe the asymptotic forms of the free energy functions.
\begin{lemma}
\label{lem:asym_FE}
Let $w^*_{XY}\in W_t$ be the true parameter, where $p(x,y|w^*_{XY})=q(x,y)$.
The free energy functions have the following asymptotic forms:
\begin{align}
F_X(n) =& \frac{\dim w}{2}\ln \frac{n}{2\pi e} +\ln\frac{\sqrt{\det I_X(w^*_{XY})}}{\varphi(w^*_{XY};\eta)} + o(1),\\
F_{XY}(n) =& \frac{\dim w}{2}\ln \frac{n}{2\pi e} +\ln\frac{\sqrt{\det I_{XY}(w^*_{XY})}}{\varphi(w^*_{XY};\eta)} + o(1),\\
F_{XY_{11}}(n) =& \frac{\dim w}{2}\ln \frac{n}{2\pi e} +\ln\frac{\sqrt{\det I_{XY_{11}}(w^*_{XY})}}{\varphi(w^*_{XY};\eta)} + o(1).
\end{align}
\end{lemma}
Based on the relations given in Eqs.~(\ref{eq:rel_n1}) and (\ref{eq:rel_n2}), we obtain the following theorem.
\begin{theorem}
\label{th:asym_Dn}
The error functions in the non-redundant case have the following asymptotic forms:
\begin{align}
D_{n1}(n) = \frac{1}{2n}\ln \det I_{XY}(w^*_{XY}) I_X(w^*_{XY})^{-1} + o\bigg(\frac{1}{n}\bigg),\\
D_{n2}(n) = \frac{1}{2n}\ln \det I_{XY_{11}}(w^*_{XY}) I_X(w^*_{XY})^{-1} + o\bigg(\frac{1}{n}\bigg).
\end{align}
\end{theorem}
This theorem shows that the leading term has the order $1/n$, and its coefficient depends on the true parameter $w^*_{XY}$.
Let us look at $D_{n1}(n)$ in order to evaluate how the true parameter affects this coefficient.
This coefficient can be rewritten as
\begin{align}
\ln \det I_X(w^*_{XY})^{-1} - \ln \det I_{XY}(w^*_{XY})^{-1}.
\end{align}
Since the inverse Fisher information matrix corresponds to the variance of the unbiased estimator,
this coefficient shows the difference between the variance of the estimator based on the incomplete dataset $X^n$ 
and that based on the complete dataset $(X^n,Y^n)$.
This is regarded as the gain in information when we obtain the latent variable $Y^n$.
For example, when the clustering is difficult, i.e., when the clusters are close together,
a partially given label $y$ considerably improves the result.
In this case, the difference between the variances will be large and so will the error.
The location of the true parameter $w^*_{XY}$ determines the degree of difficulty of the clustering task,
and the coefficient of the error reflects this difficulty through the variances of the unbiased estimators.

Lemma \ref{lem:Dn1geDn2} stated the relation between the magnitudes of the error functions.
According to Theorem \ref{th:asym_Dn}, the difference between the errors is revealed in detail in the asymptotic situation.
\begin{corollary}
The asymptotic difference between $D_{n1}(n)$ and $D_{n2}(n)$ is 
\begin{align}
D_{n1}(n) - D_{n2}(n) =& \frac{1}{2n} \ln \det I_{XY}(w^*_{XY}) I_{XY_{11}}(w^*_{XY})^{-1} + o\bigg(\frac{1}{n}\bigg).
\end{align}
\end{corollary}
Because $I_{XY}(w^*_{XY}) I_{XY_{11}}(w^*_{XY})^{-1}$ is not the unit matrix,
the coefficient of the leading term is positive; thus, this difference is of the order of $1/n$.
\subsection{Asymptotic Errors in the Redundant Case}
\label{subsec:redundant}
In the redundant case, it is known that the Fisher information matrix is not positive definite,
and the asymptotic form of the free energy is not described by that matrix.
The parameter space includes singularities, but there is an analysis method based on algebraic geometry that is able to deal effectively with singularities \cite{Watanabe01a,Watanabe09:book,Yamazaki15a}.

Let us first define some functions:
\begin{align}
H_X(w) =& \int q(x)\ln\frac{q(x)}{p(x|w)}dx,\\
H_{XY_{11}}(w) =& \int \sum_{y^{(11)}:K_t}^{L^*} q(x,y^{(11)})\ln \frac{q(x,y^{(11)})}
{\sum_{y^{(12)}=1}^{L}\sum_{y^{(2)}=1}^Lp(x,y^{(11)},y^{(12)},y^{(2)})}dx,\label{eq:def_HXY11}\\
H_{XY_1}(w) =& \int \sum_{y^{(1)}:K^*}^{L^*}  q(x,y^{(1)})\ln\frac{q(x,y^{(1)})}
{\sum_{y^{(2)}=1}^Lp(x,y^{(1)},y^{(2)}|w)}dx,\label{eq:def_HXY1}\\
H_{XY_1C}(w) =& \int \sum_{y^{(1)}:K^*}^{L^*} q(x,y^{(1)})\ln\frac{q(x,y^{(1)})}
{p(x,y^{(1)},y^{(2)}=1)}dx.
\end{align}
Instead of the Fisher information matrix,
the following zeta function plays an important role in expressing the asymptotic form of the error in the redundant case:
\begin{align}
\zeta_{A}(z) =& \int H_{A}(w)^z\varphi(w;\eta)dw,
\end{align}
where $z$ is a one-dimensional complex variable, and the symbol $A$ is replaced by $X,XY_{11},XY_1$, or $XY_1C$.
When $H_A(w)$ is an analytic function,
the poles of the zeta function are all real, negative, and rational.
Let the largest pole and the order of the corresponding zeta function
be $z=-\lambda_A$ and $m_A$, respectively;
the zeta function is 
\begin{align}
\zeta_A(z) =& \frac{f_c(z)}{(z+\lambda_A)^{m_A}} + \dots,
\end{align}
where $f_c$ is holomorphic and does not have a factor ($z+\lambda_A$).
Using this information about the pole, we can obtain the asymptotic form of the free energy function.
\begin{lemma}[Corollary 6.1 in \cite{Watanabe09:book}]
\label{lem:asym_FE_sing}
Let $H_A(w)$ be an analytic function, and let $\varphi(w;\eta)$ be analytic in its support. 
Then, the free energy function has the asymptotic form
\begin{align}
F_A(n) =& \lambda_A \ln n - (m_A-1)\ln \ln n +O(1).
\end{align}
\end{lemma}

We can then obtain the asymptotic errors from the relations given in Eqs.~(\ref{eq:rel_r1}), (\ref{eq:rel_r2}), and (\ref{eq:rel_r3}).
\begin{theorem}
\label{th:asym_Dr}
If $W_t\neq \emptyset$ and the conditions of Lemma \ref{lem:asym_FE_sing} are satisfied, 
the errors have the following asymptotic form:
\begin{align}
D_{r1}(n) =& (\lambda_{XY_1C}-\lambda_X)\frac{\ln n}{n} - (m_{XY_1C}-m_X)\frac{\ln\ln n}{n}
 +o\bigg(\frac{\ln\ln n}{n}\bigg),\\
D_{r2}(n) =& (\lambda_{XY_1}-\lambda_X)\frac{\ln n}{n} - (m_{XY_1}-m_X)\frac{\ln\ln n}{n}
 +o\bigg(\frac{\ln\ln n}{n}\bigg),\\
D_{r3}(n) =& (\lambda_{XY_{11}}-\lambda_X)\frac{\ln n}{n} -(m_{XY_{11}}-m_X)\frac{\ln \ln n}{n}
 +o\bigg(\frac{\ln\ln n}{n}\bigg).
\end{align}
\end{theorem}
The following lemma guarantees that the largest order of any of these errors is $\ln n /n$.
\begin{lemma}
\label{lem:pos_coef}
In the asymptotic forms of Theorem \ref{th:asym_Dr},
the coefficients of the leading terms on the right-hand side are all positive.
\end{lemma}
The errors in the redundant case are much larger than those in the non-redundant case.
Since this lemma is not directly derived from the asymptotic forms of the free energy functions,
we will show the proof.

{(\bf Proof of Lemma \ref{lem:pos_coef}):}
The magnitude relation $H_1(w)\le H_2(w)$
ensures that $\lambda_1\le\lambda_2$,
where $z=-\lambda_1$ and $z=-\lambda_2$ are the largest poles
of $\int H_1(w)^z\varphi(w)dw$ and $\int H_2(w)^z\varphi(w)dw$, respectively \cite{Watanabe01a,Yamazaki03a,Yamazaki10a,Watanabe09:book}.

Based on the log-sum inequality, we easily obtain
\begin{align}
H_{XY_1C}(w) > H_X(w),\\
H_{XY_1}(w) > H_X(w),\\
H_{XY_{11}}(w) > H_X(w).
\end{align}
For example,
\begin{align}
H_{XY_1C}(w) =& \int \sum_{y^{(1)}:K^*}^{L^*} q(x,y^{(1)})\ln \frac{q(x,y^{(1)})}{p(x,y^{(1)},y^{(2)}=1|w)}dx\nonumber\\
\ge& \int q(x)\ln \frac{q(x)}{p(x,y^{(2)}=1|w)}dx\nonumber\\
>& \int q(x)\ln \frac{q(x)}{p(x|w)}dx =H_X(w).
\end{align}
Therefore, $\lambda_{XY_1C}>\lambda_X$, $\lambda_{XY_1}>\lambda_X$, and $\lambda_{XY_{11}}>\lambda_X$,
which shows that the coefficients are positive.
{\bf (End of Proof)}

The differences between the errors can be derived directly.
\begin{corollary}
\label{cor:diff_sing}
The asymptotic difference between $D_{r1}(n)$ and $D_{r2}(n)$ can be expressed as
\begin{align}
D_{r1}(n)-D_{r2}(n) =& (\lambda_{XY_1C}-\lambda_{XY_1})\frac{\ln n}{n}\nonumber\\
& - (m_{XY_1C}-m_{XY_1})\frac{\ln\ln n}{n} +o\bigg(\frac{\ln\ln n}{n}\bigg),
\end{align}
and the one between $D_{r2}(n)$ and $D_{r3}(n)$ can be expressed as
\begin{align}
D_{r2}(n)-D_{r3}(n) =& (\lambda_{XY_1}-\lambda_{XY_{11}})\frac{\ln n}{n}\nonumber\\
& - (m_{XY_1}-m_{XY_{11}})\frac{\ln\ln n}{n} +o\bigg(\frac{\ln\ln n}{n}\bigg).
\end{align}
\end{corollary}
\subsection{Extension to $L>L^*$}
\label{subsec:extGenCase}
In this subsection, the extension to the case $L>L^*$ is introduced in both non-redundant and redundant cases.
The non-redundant case, where $K=K^*$ and $L>L^*$, corresponds to the one-dimensional latent variable in each dimension.
As stated in \cite{Yamazaki15a}, the redundancy of the range of variable causes singularities of the parameter space
even in the non-redundant case.
According to this study, the following lemma holds,
\begin{corollary}
\label{cor:exL}
The error functions in the non-redundant case have the following asymptotic forms;
\begin{align*}
D_{n1}(n) &= (\lambda_{XY}-\lambda_X)\frac{\ln n}{n} - (m_{XY}-m_X)\frac{\ln\ln n}{n}
 +o\bigg(\frac{\ln\ln n}{n}\bigg),\\
D_{n2}(n) &= (\lambda_{XY_{11}}-\lambda_X)\frac{\ln n}{n} - (m_{XY_{11}}-m_X)\frac{\ln\ln n}{n}
 +o\bigg(\frac{\ln\ln n}{n}\bigg).
\end{align*}
\end{corollary}
Note that the coefficients $\lambda_X$ and $\lambda_{XY_{11}}$ are different from the ones in Theorem \ref{th:asym_Dr},
since $H_X(w)$ and $H_{XY_{11}}(w)$ depend on the relation between $L$ and $L^*$.
Corollary \ref{cor:exL} can be extended to the case, where $K=K^*$
and at least one of the ranges $L_1,\dots,L_{K^*}$ is larger than the true range.
More precisely, the corollary still holds when there exists $1\le i\le K^*$ such that $L_i>L^*_i$.

In redundant case, all results in Section \ref{subsec:redundant} still hold;
the asymptotic form based on $\lambda_A$ does not change
though the function $H_A(w)$ such as Eqs. (\ref{eq:def_HXY11}), and (\ref{eq:def_HXY1})
and the value of $\lambda_A$ will change due to the extension.
Even in the case, where $L^*_1,\dots,L^*_{K^*}$ are different and at least one of them has redundancy
such as $L_i>L^*_i$, Lemma \ref{lem:asym_FE_sing} still holds.
Therefore, the asymptotic form of the error functions in Theorem \ref{th:asym_Dr}, the analysis of the largest order in Lemma \ref{lem:pos_coef},
and the asymptotic differences between the error functions in Corollary \ref{cor:diff_sing} are available even in the more general cases on $L$.
\section{Errors in Two-Layered Bayesian Networks}
\label{sec:app2BN}
In the non-redundant case, we can interpret the asymptotic forms of the errors, 
since the Fisher information matrices indicate the variance of the estimators.
However, in the redundant case, it is not straightforward to understand the meaning of the coefficients of the errors.
In this section, we provide a demonstration of a method for calculating the error in a Bayesian network,
and we ascertain the main factor determining $\lambda_A$.
For simplicity, we focus on the errors $D_{r1}(n)$ and $D_{r2}(n)$, individually and in comparison.
\subsection{Model Settings}
Let the learning model be the two-layered Bayesian network in which all the observable and hidden nodes are binary.
\begin{figure}[t]
\centering
\includegraphics[width=45mm, angle=-90]{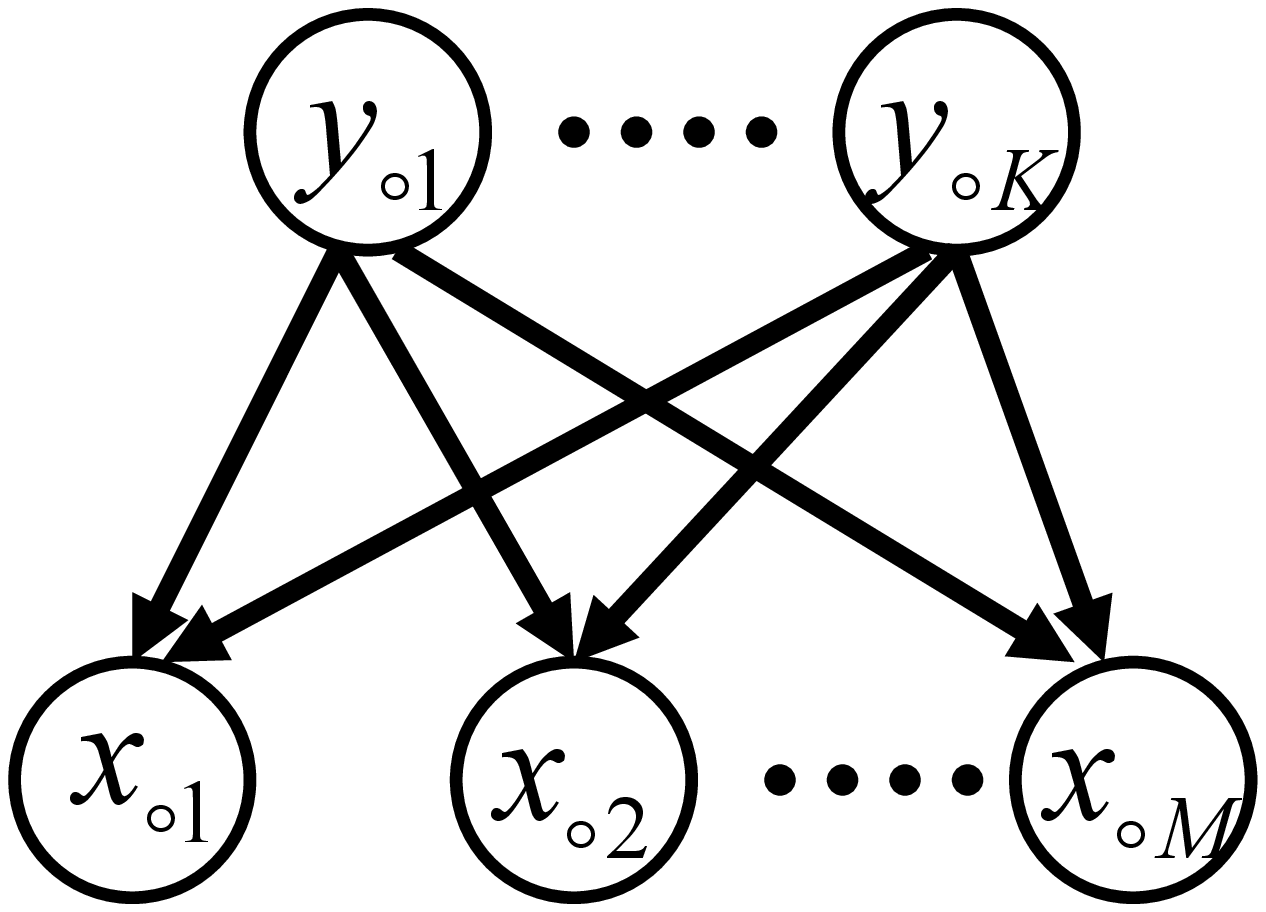}
\caption{Structure of a two-layered Bayesian network.}
\label{fig:2layeredBN}
\end{figure}
The model structure is shown in Fig. \ref{fig:2layeredBN}.
There are $M$ observable nodes directly affected by $K$ hidden ones.
Data are expressed as $x \in \{0,1\}^M$, and
the corresponding latent variables are $y \in \{0,1\}^K$.
Assume that $n$ data points are obtained.
The $i$th data point and the latent variables
are given by $x_i=(x_{i1},\dots,x_{iM})$ and $y_i=(y_{i1},\dots,y_{iK})$, respectively.
The model is formulated as
\begin{align}
p(x_i|w) =& \sum_{y_{i1}=0,1}\dots
\sum_{y_{iK}=0,1}\prod_{k=1}^K g(y_{ik},a_k)\prod_{m=1}^M g(x_{im},b_{y_im}),\label{eq:BNexpression}\\
g(j,p) =& p^{1-j}(1-p)^j,
\end{align}
where $j\in\{0,1\}$ and $0\le p \le 1$.
The parameter $w$ consists of $a_k$ and $b_{y_im}$.
The dimension is $\dim w = K + 2^KM$.

Let the prior be the product of the beta distributions:
\begin{align}
\varphi(w;\eta) =& \prod_{k=1}^K \mathrm{B}(a_k;\eta_1)\prod_{y_{i1}=0,1}\dots \prod_{y_{iK}=0,1}\prod_{m=1}^M \mathrm{B}(b_{y_im};\eta_2),
\end{align}
where $\mathrm{B}(\cdot;\eta_i)$ for $i=1,2$ is the symmetric beta function,
and $\eta=(\eta_1,\eta_2)$ is the hyperparameter.

Let the true model be a two-layered Bayesian network with $K^*$ hidden nodes.
The model has constant parameters $0<a^*_k<1$ for $1\le k \le K^*$
and $0<b^*_{y_im}<1$ for $y_i \in \{0,1\}^{K^*}$.
\subsection{Asymptotic Errors}
In the two-layered Bayesian networks, the errors of the latent variable estimation have the following behavior.
\begin{theorem}
\label{th:BNerror}
In a two-layered Bayesian network, the errors have the following upper bound: 
\begin{align}
D_{r1}(n) <& (K-K^*)\eta_1\frac{\ln n}{n} + o\bigg(\frac{\ln n}{n}\bigg),\\
D_{r2}(n) <& \min \bigg\{ \frac{(K-K^*)\eta_1}{2},2^{K-K^*-1}M \bigg\}\frac{\ln n}{n} + o\bigg(\frac{\ln n}{n}\bigg).\label{eq:D2bound}
\end{align}
Moreover, the difference has the lower bound
\begin{align}
D_{r1}(n) - D_{r2}(n) >& 
\min \bigg\{ \frac{(K-K^*)\eta_1}{2},2^{K-K^*-1}M\bigg\}\frac{\ln n}{n} + o\bigg(\frac{\ln n}{n}\bigg).\label{eq:asym_diff_D1D2}
\end{align}
In both Eqs.~(\ref{eq:D2bound}) and (\ref{eq:asym_diff_D1D2}), the coefficients of the bounds change at
\begin{align}
\eta_1=& \frac{2^{K-K^*}M}{K-K^*}.
\end{align}
\end{theorem}

From Eq.~\ref{eq:asym_diff_D1D2}, we immediately obtain the lower bound on $D_{r1}(n)$:
\begin{align}
D_{r1}(n) >& \min \bigg\{ \frac{(K-K^*)\eta_1}{2},2^{K-K^*-1}M\bigg\}\frac{\ln n}{n} + o\bigg(\frac{\ln n}{n}\bigg).
\end{align}
Let us consider the case in which $\eta_1$ is smaller than $\frac{2^{K-K^*}M}{K-K^*}$.
Combining the lower and the upper bound,
we find that
\begin{align}
\frac{(K-K^*)\eta_1}{2}\frac{\ln n}{n} + o\bigg(\frac{\ln n}{n}\bigg)
< D_{r1}(n) < (K-K^*)\eta_1 \frac{\ln n}{n} + o\bigg(\frac{\ln n}{n}\bigg),
\end{align}
which holds for arbitrary $M$.
This relation indicates that the error $D_{r1}(n)$ is essentially determined by
the factor $K-K^*$, which is the redundancy of the latent variable in the model.
\subsection{Proof of Theorem \ref{th:BNerror}}
\label{sec:proof}
Based on the relations to the free energy function,
we will show the following lemma
and then prove Theorem \ref{th:BNerror}.
\begin{lemma}
\label{lem:BNFE}
The free energy functions have the following asymptotic bounds:
\begin{align}
F_X(n) >& \frac{K^*+2^{K^*}M}{2}\ln n + o(\ln n),\\
F_{XY_1C}(n) =&\bigg\{\frac{K^*+2^{K^*}M}{2}+(K-K^*)\eta_1\bigg\}\ln n +o(\ln n),\\
F_{XY_1}(n) \le& \frac{K^*+2^{K^*}M+\min\{(K-K^*)\eta_1,2^{K-K^*}M\}}{2}\ln n +o(\ln n).
\end{align}
\end{lemma}

{\bf (Proof of Lemma \ref{lem:BNFE}):}
Since $F_X(n)$ is the free energy with respect to the probability of the observable variable $p(x|w)$,
its asymptotic properties have been clarified thoroughly (cf. \cite{Schwarz,Watanabe01a,Watanabe09:book}).
It is known that
the coefficient $2\lambda_X$ requires a value other than
the number of dimensions of the true constant parameters.
In our case, the true model has a constant parameter consisting of  $a^*_k$ and $b^*_{y_im}$,
where $1\le k \le K^*$ and $y_i\in \{1,\dots,2^{K^*}\}$.
Then, we have
\begin{align}
\lambda_X > \frac{K^*+2^{K^*}M}{2},
\end{align}
where $2^{K^*}M$ is the number of dimensions of the parameter in the true model.
Thus, we have derived the lower bound on $F_X(n)$.

Next, we consider $\zeta_{XY_1C}(z)$ for $F_{XY_1C}(n)$.
We need to calculate the integral of the zeta function
to obtain the pole.
The calculation is easy if $H_A(w)$ is in product form:
\begin{align}
\zeta_A(z) =& \int \prod_i w_i^{\alpha z}\varphi(w;\eta)dw.
\end{align}
Assuming that the prior is described by $\prod_i w_i^{\beta_i}$,
then we have the poles $z=-(\beta_i+1)/\alpha_i$,
because 
\begin{align}
\int w_i^{\alpha_iz}w_i^{\beta_i}dw_i = \frac{1}{\alpha_iz+\beta_i+1}.
\end{align}
However, $H_A(w)$ is usually in polynomial form.
One way to resolve the singularities is to use a procedure that finds a mapping that transforms $H_A(w)$
into the product form \cite{Hironaka,Watanabe01a}.
This mapping is referred to as a blow-up, and it can be used to obtain the largest pole.

The function $H_{XY_1C}(w)$ can be rewritten as
\begin{align}
H_{XY_1C}(w) =& H_{11}(w)+H_{12}(w),\\
H_{11}(w) =& \sum_{k=1}^{K^*}\sum_{j=0,1}g(j,a^*_k)\ln\frac{g(j,a^*_k)}{g(j,a_k)}\nonumber\\
&\hskip-8mm+\!\!\!\!\! \sum_{y^{(1)}:K^*}\sum_{m=1}^M\sum_x g(x_{\cdot m},b^*_{y^{(1)}m})
\ln\frac{g(x_{\cdot m},b^*_{y^{(1)}m})}{g(x_{\cdot m},b_{y^{(1)}1\dots 1m})},\\
H_{12}(w) =& -\sum_{k=K^*+1}^K \ln (1-a_k).
\end{align}
The magnitude relation $\lambda_1\le\lambda_2$ holds for $H_1(w)\le H_2(w)$,
where $z=-\lambda_1$ and $z=-\lambda_2$ are the largest poles
of $\int H_1(w)^z\varphi(w)dw$ and $\int H_2(w)^z\varphi(w)dw$, respectively (cf. the proof of Lemma \ref{lem:pos_coef}).
Moreover, $\lambda_1=\lambda_2$ when there are positive constants $c_1$ and $c_2$
such that $c_1H_1(w)\le H_2(w) \le c_2H_1(w)$.
In this sense, we use the equivalence relation $H_1(w)\equiv H_2(w)$.

Thus we have
\begin{align}
H_{XY_1C}(w) \equiv& \sum_{k=1}^{K^*} (a_k-a^*_k)^2 
+ \sum_{y^{(1)}:K^*}\sum_{m=1}^M (b_{y^{(1)}1\dots 1m}-b^*_{y^{(1)}m})^2 + \sum_{k=K^*+1}^K a_k.
\end{align}
We will denote the right-hand side as $H_{13}(w)$.
We define the blow-up $\Psi_1:\bar{w}=(u,v)\rightarrow w$ as follows:
\begin{align}
u_1 =& a_1-a^*_1 \\
u_k u_1 =& a_k-a^*_k \;\; (1<k\le K^*)\\
u_k u_1^2 =& a_k \;\; (K^*<k\le K)\\
v_{y^{(1)}1\dots 1 m} u_1=& b_{y^{(1)}1\dots 1 m} - b^*_{y^{(1)}m}.
\end{align}
We focus on the factor $u_1$ in the zeta function
\begin{align}
\int H_{13}(\Psi_1(\bar{w}))^z\varphi(\Psi_1(\bar{w});\eta)|\Psi_1|d\bar{w},
\end{align}
where $|\Psi_1|$ is a Jacobian.
Consider the exponential part of $u_1$.
In the factor $H_{13}(\Psi_1(\bar{w}))^z$
and $K^*+2^{K^*}M+2(K-K^*)\eta_1-1$ in the beta prior $\varphi(\Psi_1(\bar{w});\eta)$, we have $2z$, which results in the pole
\begin{align}
z=-\frac{K^*+2^{K^*}M}{2}-(K-K^*)\eta_1.
\end{align}
Even if we focus on different factors $u_i$ for $i=2,\dots,K^*$,
we will obtain the same pole; this is due to the symmetry of $a_2,\dots,a_{K^*}$.
Thus, this is the largest pole among the transformed coordinates of $w$.
According to the equivalence relation, the pole has the value $-\lambda_{XY_1C}$.

For the bounds on $F_{XY_1}(n)$,
we analyze $H_{XY_1}(w)$.
According to \cite{Yamazaki10a,Watanabe09:book},
this is equivalent to the squared-error form for discrete models.
Applying this property to our model, we obtain
\begin{align}
H_{XY_1}(w) \equiv& \sum_x \sum_{y^{(1)}}\{ \sum_{y^(2)}p(x,y^{(1)},y^{(2)}) - q(x,y^{(1)}) \}^2.
\end{align}

Let us define the blow-up $\Psi_2:\bar{w}\rightarrow w$ such that
\begin{align}
u_1 =& a_1-a^*_1,\label{eq:Psi2begin}\\
u_ku_1 =& a_k-a^*_k \;\; (1<k\le K^*)\\
u_ku_1 =& a_k \;\; (K^*<k\le K)\\
v_{y^{(1)}1\dots 1m}u_1 =& b_{y^{(1)}1\dots 1m}-b^*_{y^{(1)}m}\label{eq:Psi2end}.
\end{align}
Then, there is a positive constant $c_3$ such that
\begin{align}
H_{XY_1}(\Psi_2(\bar{w})) \le c_3 u_1^2
\end{align}
in the support of $\varphi(\Psi_2(\bar{w});\eta)$.
In the zeta function
\begin{align}
\int u_1^{2z}\varphi(\Psi_2(\bar{w});\eta)|\Psi_2|d\bar{w},
\end{align}
$u_1$ has the exponential part $2z+K^*+2^{K^*}M+(K-K^*)\eta_1-1$,
which results in the pole $z=-\mu_1=-(K^*+2^{K^*}M+(K-K^*)\eta_1)/2$.

Let us define the blow-up $\Psi_3:\bar{w}\rightarrow w$ such that
\begin{align}
u_1 =& a_1-a^*_1,\label{eq:Psi3begin}\\
u_ku_1 =& a_k-a^*_k \;\; (1<k\le K^*)\\
v_{y^{(1)}y^{(2)}m}u_1 =& b_{y^{(1)}y^{(2)}m}-b^*_{y^{(1)}m}.\label{eq:Psi3end}
\end{align}
Then there is a positive constant $c_4$ such that
\begin{align}
H_{XY_1}(\Psi_3(\bar{w})) \le c_4 u_1^2
\end{align}
in the support of $\varphi(\Psi_3(\bar{w});\eta)$.
In the zeta function
\begin{align}
\int u_1^{2z}\varphi(\Psi_3(\bar{w});\eta)|\Psi_3|d\bar{w},
\end{align}
$u_1$ has the exponential part $2z+K^*+2^{K^*}M+2^{K-K^*}M-1$,
which results in the pole $z=-\mu_2=-(K^*+2^{K^*}M+2^{K-K^*}M)/2$.

Then, it holds that $\lambda_{XY_1}\le \min\{\mu_1,\mu_2\}$,
which proves the upper bound on $F_{XY_1}(n)$.
{\bf (End of Proof)}

{\bf (Proof of Theorem \ref{th:BNerror}):}
Using Eqs.~(\ref{eq:rel_r1}) and (\ref{eq:rel_r2}),
the asymptotic bounds in Lemma \ref{lem:BNFE} immediately show
the bounds on $D_{r1}(n)$ and $D_{r2}(n)$.
Their difference is written as
\begin{align}
D_{r1}(n)-D_{r2}(n) =& \frac{1}{n}\bigg\{F_{XY_1C}(n)-F_{XY_1}(n)\bigg\},
\end{align}
and its lower bound is also derived from the asymptotic properties of the free energy functions.
{\bf (End of Proof)}
\section{Discussion}
\label{sec:Dis}
In this section, we first explain the phase and its transition in the free energy function,
which indicates how the redundant dimensions of the latent variable are used and affects the accuracy of the estimation.
Then, we discuss the symmetry of dimension in the latent variable.
%
\subsection{Phase Transition and Its Effect on the Estimation}
Here, we introduce the phase transition of the free energy function
and discuss parameter subspaces used to determine the phases.
The relation between the phases and the subspaces provides 
an interpretation of the coefficient of the dominant term in the energy function.

Previous studies, such as \cite{Yamazaki13a}, have shown that at some points, there is a change in the shape of the free energy function with respect to the hyperparameter.
This change is referred to as a phase transition, and the point at which the change occurs is the phase transition point.
For example, in Lemma \ref{lem:BNFE}, the bound on $F_{XY_1}(n)$ has two phases:
\begin{align}
\frac{K^*+2^{K^*}M+(K-K^*)\eta_1}{2}\ln n + o(\ln n) \;\;\; (\eta_1<\eta_t),\label{eq:1st_bound}\\
\frac{K^*+2^{K^*}M+2^{K-K^*}M}{2}\ln n + o(\ln n) \;\;\; (\eta_1\ge \eta_t),\label{eq:2nd_bound}
\end{align}
and the transition point is 
\begin{align}
\eta_t =& \frac{2^{K-K^*}M}{K-K^*}.
\end{align}
The difference between the coefficients in Eqs.~(\ref{eq:1st_bound}) and those in (\ref{eq:2nd_bound}) is
based on the blow-up mappings $\Psi_2(\bar{w})$ and $\Psi_3(\bar{w})$ in the proof of Lemma \ref{lem:BNFE}
defined by Eqs.~(\ref{eq:Psi2begin})-(\ref{eq:Psi2end}) and (\ref{eq:Psi3begin})-(\ref{eq:Psi3end}), respectively.
These mappings show the subspaces of the parameter space,
which provide dominant values in the calculation of the integral of the zeta function $\zeta_{XY_1}(z)$
and determine the locations of the poles.
In the redundant case, the true parameter $w^*_{XY}$ is expressed as not an isolated point
but a solution set of some equations such as Eqs.~(\ref{eq:Psi2begin})-(\ref{eq:Psi2end}).

Since the subspace generally appears in the analysis of the hierarchical models
and it is not straightforward to elucidate its complicated structure,
let us consider a simple case, where the model is the two-layered Bayesian network shown in Fig.~\ref{fig:simpleBN},
and the true distribution has a one-dimensional latent variable; thus, we have $K^*=1$ and $K=2$.
The model and the true distribution are given by
\begin{align}
p(x|w) =& \sum_{y_1=0,1}\sum_{y_2=0,1} a_{1y_1}a_{2y_2}\prod_{m=1}^M b_{y_1y_2m}^{(1-x_m)}(1-b_{y_1y_2m})^{x_m},\label{eq:p_model_rep}\\
q(x) =& \sum_{y_1=0,1} a^*_{1y_1}\prod_{m=1}^M b^{*(1-x_m)}_{y_1m}(1-b^*_{y_1m})^{x_m},
\end{align}
where Eq.~(\ref{eq:p_model_rep}) is the same expression as Eq.~(\ref{eq:def_p_simpleBN}).
The necessary and redundant dimensions are expressed as $Y_1=\{y_1\}$ and $Y_2=\{y_2\}$, respectively.
The distribution of the estimation is given by
\begin{align}
p(Y_1|X^n) =& \sum_{y_2=0,1} p(Y_1,Y_2|X^n).
\end{align}
We can find at least two ways to express the true distribution while satisfying $p(Y_1|X^n)=q(Y_1|X^n)$:

\begin{description}
\item[(P1)] $a_{20}=0$,
\item[(P2)] $b_{y_1y_2m}=b^*_{y_1m}$ for any $y_2$.
\end{description}
\begin{figure}[t]
\def\@captype{table}
\begin{minipage}[c]{.5\textwidth}
\begin{tabular}{c||c|c}
$y_1$ & 0 & 1\\
\hline
& $a^*_{10}$ & $1-a^*_{10}$
\end{tabular}
\\
\begin{tabular}{c}
\\
\end{tabular}
\\
\begin{tabular}{c||c|c}
$y_2$ & 0 & 1 \\
\hline
& 0 & 1
\end{tabular}
\end{minipage}
\begin{minipage}[c]{.5\textwidth}
\begin{tabular}{c||c|c|c|c}
$x_m$ & $y_1$ & $y_2$ & 0 & 1 \\
\hline
& 0 & 0 & FV & --- \\
& 0 & 1 & $b^*_{0m}$ & $1-b^*_{0m}$ \\
& 1 & 0 & FV & --- \\
& 1 & 1 & $b^*_{1m}$ & $1-b^*_{1m}$
\end{tabular}
\end{minipage}
\caption{Parameter of {\bf (P1)}: the CPT of the nodes $y_1$ and $y_2$ (left panel)
and the CPT of the node $x_m$ (right panel); FV indicates a free variable.}
\label{fig:P1}
\end{figure}
The first way {\bf (P1)} directly eliminates the effect of $y_2$, since $y_2$ is always unity,
and the actual latent variable is reduced to $y_1$ in Eq.~(\ref{eq:p_model_rep}).
The conditional probability tables (CPTs) are presented in Fig.~\ref{fig:P1}. 
We refer to this as the \emph{eliminating} way.
In the CPT of $x_m$, FV indicates a free variable.
For example, in the first row of the CPT,
$p(x_m=0|y_1=0,y_2=0,w)=b_{00m}$ is an FV and may take any value in the range $[0,1]$;
$p(x_m=1|y_1=0,y_2=0,w)=1-b_{00m}$ is omitted from the CPT.

\begin{figure}[t]
\def\@captype{table}
\begin{minipage}[c]{.5\textwidth}
\begin{tabular}{c||c|c}
$y_1$ & 0 & 1\\
\hline
& $a^*_{10}$ & $1-a^*_{10}$
\end{tabular}
\\
\begin{tabular}{c}
\\
\end{tabular}
\\
\begin{tabular}{c||c|c}
$y_2$ & 0 & 1 \\
\hline
& FV & ---
\end{tabular}
\end{minipage}
\begin{minipage}[c]{.5\textwidth}
\begin{tabular}{c||c|c|c|c}
$x_m$ & $y_1$ & $y_2$ & 0 & 1 \\
\hline
& 0 & 0 & $b^*_{0m}$ & $1-b^*_{0m}$ \\
& 0 & 1 & $b^*_{0m}$ & $1-b^*_{0m}$ \\
& 1 & 0 & $b^*_{1m}$ & $1-b^*_{1m}$ \\
& 1 & 1 & $b^*_{1m}$ & $1-b^*_{1m}$
\end{tabular}
\end{minipage}
\caption{Parameter of {\bf (P2)}: the CPT of the nodes $y_1$ and $y_2$ (left panel)
and the CPT of the node $x_m$ (right panel); FV indicates a free variable.}
\label{fig:P2}
\end{figure}
The second way {\bf (P2)} determines the value of $p(x|y_1,y_2,w)$, which depends only on $y_1$. The CPTs are shown in
Fig.~\ref{fig:P2}.
We refer to this as the \emph{replicating} way, since the rows in the CPT of $x_m$ are replicated
so that we can ignore the value of $y_2$.
Naturally, $p(y_2=0|w)=a_{20}$ is the free variable.

Let us describe the subspaces for {\bf (P1)} and {\bf (P2)}.
According to the CPTs, the eliminating way can be expressed as
\begin{align}
a_{10} =& a^*_{10},\\
a_{20} =& 0,\\
b_{y_11m} =& b^*_{y_1m},
\end{align}
which corresponds to the subspace defined by the blow-up $\Psi_2$ in Section \ref{sec:proof}.
The replicating way can be expressed as
\begin{align}
a_{10} =& a^*_{10},\\
b_{y_1y_2m} =& b^*_{y_1m},
\end{align}
which corresponds to the subspace defined by $\Psi_3$.
Based on the proof of Lemma \ref{lem:BNFE},
$\Psi_2$ and $\Psi_3$ provide the poles of the zeta function, and
the coefficients in Eqs.~(\ref{eq:1st_bound}) and (\ref{eq:2nd_bound}) are then derived.
Therefore, the first phase, represented by Eq.~(\ref{eq:1st_bound}),
implies that the eliminating way can be used to express the true distribution,
while the second phase, represented by Eq.~(\ref{eq:2nd_bound}), implies the replicating way can be used.

On the other hand, there is no transition point in $F_{XY_1C}(n)$.
Its definition includes the factor $p(x_i,y^{(1)}_i,y^{(2)}_i=1|w)$,
where the redundant dimension must have a fixed value, and
the other values of $y^{(2)}_i$ do not have any effect on the observation $x_i$.
Then, the probabilities for $y^{(2)}_i \neq 1$ must be zero,
which corresponds to {\bf (P1)}.
This is why only the eliminating way can be used to express the true distribution,
and $F_{XY_1C}(n)$ does not have different phases.

Now we discuss the effect of the phase transition on the estimation of the latent variable.
According to Theorem \ref{th:BNerror}, the bound of the error $D_{r2}(n)$ depends on the phase;
\begin{align}
D_{r2}(n) <& \frac{(K-K^*)\eta_1}{2}\frac{\ln n}{n} +o\bigg(\frac{\ln n}{n}\bigg) \;\;\; (\eta_1 < \eta_t),\\
D_{r2}(n) <& \frac{2^{K-K^*}M}{2}\frac{\ln n}{n} +o\bigg(\frac{\ln n}{n}\bigg) \;\;\; (\eta_1 \ge \eta_t).
\end{align}
Considering the way to express the true distribution,
we find that the upper expression is derived from the eliminating way
while the lower one is from the replicating way.
Therefore, {\bf (P2)} makes the error smaller than {\bf (P1)} in $\eta_1\ge \eta_t$,
that is, to minimize the error requires not only to express the true distribution
but also how to express it according to the value of the hyperparameter.
\subsection{Symmetry of Latent Variable}
\label{sec:Dis_symmetry}
In this subsection, we consider the symmetry of the latent variable
and its effect on the accuracy.
As mentioned in the previous study,
the label values have symmetry \cite{Yamazaki15a}.
For example, the labels of the Gaussian mixture model in Eq.(\ref{eq:GM}) are determined
by the locations of the components and swapping the components essentially shows the same clustering result;
the label probabilities of $w=\{a^*_1,b^*_1,b^*_2\}$ are expressed as
\begin{align}
p(x,y=1|w) =& a^*\mathcal{N}(x|b^*_1),\\
p(x,y=2|w) =& (1-a^*)\mathcal{N}(x|b^*_2)
\end{align}
and the ones of $w=\{1-a^*_1,b^*_2,b^*_1\}$ are
\begin{align}
p(x,y=1|w) =& (1-a^*)\mathcal{N}(x|b^*_2),\\
p(x,y=2|w) =& a^*\mathcal{N}(x|b^*_1).
\end{align}
They are equivalent when the labels $1$ and $2$ are swapped.

The multidimensional case, that the present paper focuses on, has two types of symmetry:
the label values and the dimension of the latent value.
In the simple two-layered binary Bayesian network defined by Eq.~(\ref{eq:def_p_simpleBN}),
the label probability with the parameter
\begin{align}
w = \{a^*_{01},a^*_{02},b^*_{00},b^*_{01},b^*_{10},b^*_{11}\}
\end{align}
and the one with
\begin{align}
w = \{ a^*_{02}, a^*_{01}, b^*_{00},b^*_{10},b^*_{01},b^*_{11}\}
\end{align}
are equivalent, 
where $y_1$ and $y_2$ have symmetric relation.
This is the symmetry of the dimension.
Note that the symmetry of the label value also exists in each dimension.
When $L_1=\dots =L_K=L$, there are $K!\times (L!)^K$ symmetric assignments of the labels
since permutations of the dimension and the label value are $K!$ and $(L!)^K$, respectively.

Let us now consider the effect of the symmetry on the definitions of the error functions.
Since the symmetric assignments should have equivalent probabilities,
the estimated value $p(Y^n|X^n)$ has also $K!\times (L!)^K$ symmetric structure.
The previous study \cite{Yamazaki15a} has shown 
that the redundancy of the label range ($L>L^*$) does not adversely affect the accuracy in the case $K=K^*=1$.
In the rest of this section,
we will extend this result to the multidimensional case, where there is the range redundancy or the dimension redundancy.
Section \ref{sec:asymmetric_evaluation} explains that the error functions asymmetrically evaluate the estimation results
when there is redundancy of the latent variables and the symmetry of the estimated distribution is different from that of the true distribution.
Section \ref{sec:local_area} shows that the asymptotic forms of the error functions are determined by the local parameter area, which is the neighborhood of the true parameter.
Then, in Section \ref{sec:effect_accuracy}, 
we will confirm that the asymmetric evaluation does not make the error functions biased and their convergence rates are expressed as the forms of Theorem \ref{th:asym_Dn} and Corollary \ref{cor:exL}.
We mainly consider the general case $K>K^*$ and $L>L^*$ in $D_{r1}(n)$, which has the largest difference of the symmetry between the estimated and the true distributions,
since we can show the similar results on the other error functions.
\subsubsection{Asymmetric evaluation of the error functions with redundancy}
\label{sec:asymmetric_evaluation}
When there is the redundancy of the range or the dimension of the latent variables,
the error functions define how the true distribution should be expressed.
The error functions $D_{n1}(n)$ and $D_{n2}(n)$ limit the summation from $L$ to $L^*$,
which implicitly assumes that the first $L^*$ values estimates the labels of the true distribution $y^{(1)}_{ik}=1,\dots,L^*$ in this order.
For the dimension redundancy, $D_{r1}(n)$ and $D_{r2}(n)$ limit the summations to $Y_1^n$,
which assumes that the first $K^*$ dimension in $y$ estimates the true distribution
and the rest of $y$ is the redundant part.
These definitions break the symmetry in the estimated distribution;
some parameters essentially indicating the true distribution are not regarded as the true parameter.
For example, there are two parameters $w^*_{XY_1Y_2=1}$ and $w^*_{XY_1Y_2=2}$ defined by
\begin{align}
\prod_{i=1}^n q(x_i,y^{(1)}_i) =& \prod_{i=1}^n p(x_i,y^{(1)}_i|y^{(2)}_i=1,w^*_{XY_1Y_2=1})\label{eq:true_p1}\\
\prod_{i=1}^n q(x_i,y^{(1)}_i) =& \prod_{i=1}^n p(x_i,y^{(1)}_i|y^{(2)}_i=2,w^*_{XY_1Y_2=2}),
\end{align}
respectively.
In the eliminating way, either $w^*_{XY_1Y_2=1}$ or $w^*_{XY_1Y_2=2}$ can express the true distribution under the condition $y^{(2)}=1$ or $y^{(2)}=2$.
Only the former one $w^*_{XY_1Y_2=1}$ is regarded as the true parameter in $D_{r1}(n)$ because the error function has the condition $y^{(2)}=1$.
So we find that, when there is redundancy $L>L^*$ or $K>K^*$,
the error functions asymmetrically evaluate the estimation results according to their definitions.

\subsubsection{The asymptotic errors determined by the local parameter area}
\label{sec:local_area}
For the eliminating way, the number of the symmetric assignments of the labels is calculated as
\begin{align}
C_r =& \frac{K!}{(K-K^*)!}\times \bigg(\frac{L!}{(L-L^*)!}\bigg)^{K^*} \times L^{K-K^*},
\end{align}
where $\frac{K!}{(K-K^*)!}\times \big(\frac{L!}{(L-L^*)!}\big)^{K^*}$ is the corresponding part to the true distribution
and $L^{K-K^*}$ is the redundant one with fixed values.
The true parameter $w^*_{XY_1Y_2=1}$ provides the assignment, 
where the first $K^*$ dimension has the same as the true distribution and the remaining dimension is fixed as one.
Then, the parameter space also has symmetric structure;
the parameter space is divided into $C_r$ localized areas.
Let the set of these areas be denoted by $\Sigma_W$.
The estimated probability $p(Y_1^n,Y_2^n|X^n)$ is written as
\begin{align}
p(Y_1^n,Y_2^n|X^n) \propto& p(X^n,Y_1^n,Y_2^n) \nonumber\\
=& \int \prod_{i=1}^n p(x_i,y^{(1)}_i,y^{(2)}_i|w)\varphi(w|\eta)dw \nonumber\\
=& \sum_{A_W \in \Sigma_W} \exp \bigg\{ \ln \int_{A_W} \prod_{i=1}^n p(x_i,y^{(1)}_i,y^{(2)}_i|w)\varphi(w|\eta)dw \bigg\}.
\end{align}
Assume that the assignment $\bar{Y}^n=(\bar{Y}_1^n,\bar{Y}_2^n=1)$ is obtained from the model with the true parameter $w^*_{XY_1Y_2=1}$
and $\bar{A}$ is its area.
When the area is focused, we also use $w^*_{\bar{A}}$ for the true parameter.
Recall that the asymptotic form of $F_{XY}(n)$ is derived from the calculation of the integral
over the localized area determined by the blow-up mapping such as $\Phi_1$.
By using the same coefficient such as $\lambda_{XY_1C}$,
the asymptotic form of $p(X^n,\bar{Y}_1^n,\bar{Y}_2^n=1)$ is expressed as
\begin{align}
p(X^n,\bar{Y}_1^n,\bar{Y}_2^n=1) =& 
\exp \bigg\{ \ln \int_{\bar{A}} \prod_{i=1}^n p(x_i,y^{(1)}_i,y^{(2)}_i=1|w)\varphi(w|\eta)dw \bigg\}\nonumber\\
& + \sum_{A_W \in \Sigma_W \setminus \{\bar{A}\}} \exp \bigg\{ \ln \int_{A_W} \prod_{i=1}^n p(x_i,y^{(1)}_i,y^{(2)}_i=1|w)\varphi(w|\eta)dw \bigg\}\nonumber\\
=& \exp\bigg\{ -nS_{w^*_{\bar{A}}}(X^n,\bar{Y}_1^n,\bar{Y}_2^n=1) - \lambda_{XY_1C}\ln n +O_p(\ln\ln n)\bigg\}\nonumber\\
& + \sum_{A_W \in \Sigma_W \setminus \{\bar{A}\}} \exp \bigg\{ \ln \int_{A_W} \prod_{i=1}^n p(x_i,y^{(1)}_i,y^{(2)}_i=1|w)\varphi(w|\eta)dw \bigg\},
\label{eq:asym_pbar}
\end{align}
where the empirical entropy is defined by
\begin{align}
S_w(X^n,Y_1^n,Y_2^n) =& -\frac{1}{n}\sum_{i=1}^n \ln p(x_i,y^{(1)}_i,y^{(2)}_i|w).
\end{align}
This asymptotic form shows that the dominant term of $-\ln p(X^n,Y_1^n,Y_2^n=1)$ is the entropy term
with respect to the true parameter such as $w^*_{\bar{A}}$.
The empirical entropy is always larger than that of the different parameter areas.
For example, though the assignment $Y^n=(\bar{Y}_1^n,Y_2=2)$ is correct in terms of the eliminating way,
it makes the entropy term larger in the area $\bar{A}$;
\begin{align}
S_{w^*_{\bar{A}}}(X^n,\bar{Y}_1^n,Y_2^n=2) >& S_{w^*_{\bar{A}}}(X^n,\bar{Y}_1^n,\bar{Y}_2^n=1).
\end{align}
Then, the asymptotic form is written as
\begin{align}
-\ln p(X^n,\bar{Y}_1^n,\bar{Y}_2^n=1) =& 
nS_{w^*_{\bar{A}}}(X^n,\bar{Y}_1^n,\bar{Y}_2^n=1) + \lambda_{XY_1C}\ln n +O_p(\ln\ln n). \label{eq:asym_ln_pbar}
\end{align}
because the second term of the last expression in Eq.\ref{eq:asym_pbar},
which is the integral on the different areas $\Sigma_W \setminus \{\bar{A}\}$,
is much smaller than the first term and does not appear in the leading terms.
In other words, this integral works as the operator selecting the true parameter to minimize the entropy term.
The asymptotic form is determined by the integral on the local area, which is the neighborhood of the true parameter.

\subsubsection{Effect on the accuracy}
\label{sec:effect_accuracy}
We now introduce the relation between the entropy factors $S_{w^*_{\bar{A}}}(X^n,\bar{Y}_1^n,\bar{Y}_2^n=1)$ and $S_{XY}$.
The error function is divided into two terms;
\begin{align}
D_{r1}(n) =& \frac{1}{n}E_{X^nY_1^n}\bigg[ \ln \frac{q(X^n,Y_1^n)}{p(X^n,Y_1^n,Y_2^n=1)}\bigg]
- \frac{1}{n}E_{X^n}\bigg[\ln \frac{q(X^n)}{Z(X^n)}\bigg]. \label{eq:Dr1_div_KL}
\end{align}
Because the second term does not include $Y^n$,
we focus on the first term.

The true distribution has $C_q$ symmetric areas of $Y^n_1$, where
\begin{align}
C_q =& K^*! \times (L^*!)^{K^*}.
\end{align}

Since the estimated result has $C_r$ symmetric areas,
\begin{align}
\prod_{i=1}^n p(x_i,y^{(1)}_i,y^{(2)}_i=1|w^*_{XY_1Y_2=1}) =& \frac{C_q}{C_r}\prod_{i=1}^n q(x_i,y^{(1)}_i),\label{eq:rel_p_q}
\end{align}
where the redundancy makes the estimated joint probability discounted due to the larger symmetric areas.
By considering the relation Eq.\ref{eq:true_p1}, this implies that
\begin{align}
\prod_{i=1}^n p(y^{(2)}_i=1|w^*_{XY_1Y_2=1}) =& \frac{C_q}{C_r}\nonumber\\
=& \bigg(\frac{K!}{(K-K^*)!K^*!}\bigg)^{-1}\times \bigg(\frac{L!}{(L-L^*)!L^*!}\bigg)^{-K^*} \times \big(L^{K-K^*}\big)^{-1}.
\end{align}
where the first factor $(\frac{K!}{(K-K^*)!K^*!})^{-1}$ is the probability to select $K^*$ dimension from $K$,
the second one $(\frac{L!}{(L-L^*)!L^*!})^{-K^*}$ is the probability to select $L^*$ range from $L$ in $K^*$ dimension,
and the third one $(L^{K-K^*})^{-1}$ is the probability to fix the values as $Y_2^n=1$.

Based on Eq. \ref{eq:rel_p_q}, $S_{w^*_{XY_1Y_2=1}}(X^n,Y_1^n,Y_2^n=1)$ is written as
\begin{align}
S_{w^*_{XY_1Y_2=1}}(X^n,Y_1^n,Y_2^n=1) =& -\frac{1}{n}\sum_{i=1}^n \ln p(x_i,y^{(1)}_i,y^{(2)}_i=1|w^*_{XY_1Y_2=1})\nonumber\\
=& -\frac{1}{n}\sum_{i=1}^n \ln q(x_i,y^{(1)}_i) + \frac{1}{n}\ln\frac{C_r}{C_q}\nonumber\\
=& -\ln q(X^n,Y_1^n) + \frac{1}{n}\ln\frac{C_r}{C_q}.
\end{align}
The first term of Eq.\ref{eq:Dr1_div_KL} is then rewritten as
\begin{align}
\frac{1}{n}E_{X^nY_1^n}\bigg[ \ln \frac{q(X^n,Y_1^n)}{p(X^n,Y_1^n,Y_2^n=1)}\bigg]
=& - S_{XY} + E_{X^nY_1^n}\bigg[S_{w^*_{XY_1Y_2=1}}(X^n,Y_1^n,Y_2^n=1)\bigg] \nonumber\\
&+ \lambda_{XY_1C}\frac{\ln n}{n} + O\bigg(\frac{\ln\ln n}{n}\bigg)\nonumber\\
=& - S_{XY} + \frac{1}{n}E_{X^nY_1^n}\bigg[ -\ln q(X^n,Y^n_1) + \ln\frac{C_r}{C_q} \bigg] \nonumber\\
&+ \lambda_{XY_1C}\frac{\ln n}{n} + O\bigg(\frac{\ln\ln n}{n}\bigg)\nonumber\\
=& - S_{XY} + S_{XY} + \frac{1}{n}\ln\frac{C_r}{C_q} \nonumber\\
&+ \lambda_{XY_1C}\frac{\ln n}{n} + O\bigg(\frac{\ln\ln n}{n}\bigg)\nonumber\\
=& \lambda_{XY_1C}\frac{\ln n}{n} + O\bigg(\frac{\ln\ln n}{n}\bigg).
\end{align}
Combining the asymptotic form of the second term of Eq.\ref{eq:Dr1_div_KL},
we find that the error converges to zero and its convergence rate is the one shown in Theorem \ref{th:asym_Dr}.
We can easily prove that the other error functions are unbiased,
replacing the constant $C_r$ with
\begin{align}
C_r =& \frac{K!}{(K-K^*)!} \times \bigg( \frac{L!}{(L-L^*)!} \bigg)^{K^*} \label{eq:Cr_Dr2}
\end{align}
for $D_{r2}(n)$, and with
\begin{align}
C_r =& \bigg( \frac{L!}{(L-L^*)!} \bigg)^{K^*} \label{eq:Cr_Dn12}
\end{align}
for $D_{n1}(n)$ and $D_{n2}(n)$ in the case $K=K^*$ and $L>L^*$.
Due to the marginalization on $Y_2^n$ in $D_{r2}(n)$,
Eq.~\ref{eq:Cr_Dr2} does not have the fixed value factor compared to the expression for $D_{r1}(n)$.
When there is no redundancy on the dimension $K=K^*$, $C_r$ in Eq.\ref{eq:Cr_Dn12} consists of the single factor to select the range.
Note that the error functions $D_{n1}(n)$ and $D_{n2}(n)$ do not have the difference on the symmetry when $K=K^*$ and $L=L^*$, that is $C_r=C_q$,
and the estimated probability of the symmetric areas has the equivalent value to the true distribution.
Thus, the symmetric estimation and the limitation of the true assignment defined by the error functions do not change the asymptotic forms of our results.
\section{Conclusions}
\label{sec:Conc}
The present paper formulated the error functions of the Bayesian estimation of  multidimensional latent variables,
and derived their asymptotic forms.
According to the number of the dimensions of the latent variables,
there are redundant and non-redundant cases.
For the asymptotic analysis, the Fisher information matrices play an important role in the non-redundant case
while the zeta functions of the algebraic geometrical method do in the redundant case.

Precise calculation of the coefficient $\lambda_A$ is necessary
to clarify the dominant order in the redundant case, and as seen in Section \ref{sec:app2BN},
this calculation is complex.
Even in a simpler structure, such as a naive Bayesian network or a tree model,
an additional mathematical technique is required \cite{Rusakov,Zwiernik11}.
The exact form of $\lambda_X$ can be derived for certain limited models,
and the algebraic geometrical approach is still being developed (e.g., \cite{Yamazaki10a,Lin2010}).
It is an important goal to obtain
a detailed analysis of these coefficients.
\section*{Acknowledgements}
This research was partially supported by a research grant by the Support Center for Advanced Telecommunications Technology Research Foundation
and by KAKENHI 15K00299.

\bibliography{LearningTheory}
\bibliographystyle{plain}
\end{document}